\pdfoutput=1

\documentclass[11pt]{article}

\usepackage[preprint]{acl}

\usepackage{times}
\usepackage{latexsym}
\usepackage{amssymb}
\usepackage{amsmath}
\usepackage{booktabs}
\usepackage{multirow}
\usepackage{float}
\usepackage{multicol}
\usepackage{color}

\usepackage{colortbl}

\usepackage[T1]{fontenc}

\usepackage[utf8]{inputenc}

\usepackage{microtype}

\usepackage{inconsolata}

\usepackage{graphicx}

\title{AdaFV: Rethinking of Visual-Language alignment for VLM acceleration}

\author{
  \textbf{Jiayi Han\textsuperscript{1,2}},
  \textbf{Liang Du\textsuperscript{3}},
  \textbf{Yiwen Wu\textsuperscript{1,2}},
  \textbf{Xiangguo Zhou\textsuperscript{1,2}},\\
  \textbf{Hongwei Du\textsuperscript{1,2}},
  \textbf{Weibo Zheng\textsuperscript{1,2}}
\\
  \textsuperscript{1}Inspur Genersoft Co. Ltd., Inspur Group Co. Ltd.\\
  \textsuperscript{2}Shandong Key Laboratory of Automated Complex Network Software Construction\\
  \textsuperscript{3}Interactive Entertainment Group, Tencent Inc.\\
\\
  \small{
    \textbf{Correspondence:} \href{hanjiayi@inspur.com}{hanjiayi@inspur.com}
  }
}

\begin{document}
\maketitle
\begin{abstract}
The success of Vision-Language Models (VLMs) often relies on high-resolution schemes that preserve image details. However, high-resolution image input results in redundant visual tokens, significantly reducing VLM efficiency. Various research efforts have aimed to reduce visual tokens by filtering out uninformative ones or aggregating their information to enhance VLM efficiency without incurring additional training costs. Some methods propose reducing visual tokens based on VLMs' self-attention, which can be biased and lead to inaccurate responses. Token reduction approaches that solely rely on visual cues are text-agnostic and fail to focus on areas most relevant to the query, particularly when the queried objects are non-salient in the image. In this work, we first conduct experiments to demonstrate that original text embeddings align with visual tokens without bias towards less important visual tokens. We then propose a self-adaptive cross-modality attention mixture mechanism that dynamically leverages the effectiveness of visual saliency and text-to-image similarity in the pre-LLM layers to select informative visual tokens. Extensive experiments show that the proposed approach achieves state-of-the-art training-free VLM acceleration performance, especially when the reduction rate is significantly high.
\end{abstract}

\section{Introduction}
In recent years, vision-language models (VLMs) have demonstrated exceptional performance in various visual-grounded tasks, including image captioning, visual question answering, and layout understanding. Despite their impressive achievements, the computational cost associated with VLMs remains a significant challenge for practical deployment. A key factor contributing to this high computational cost is the large number of visual tokens. For instance, LLaVA-NEXT models \cite{liu2024llavanext} utilize 2880 visual tokens for single-image tasks, while modern VLMs like InternVL-2.5 \cite{chen2024expanding} employ over 8,000 visual tokens per task.
To address this issue, numerous research efforts have focused on pruning redundant visual tokens to accelerate VLMs without additional training. For example, FastV \cite{chen2024image} observes that the distribution of attention weights among visual tokens tends to cluster and ranks these tokens based on their attention weights, retaining only the top-ranked tokens. Similarly, SparseVLM \cite{zhang2024sparsevlm} identifies text tokens highly relevant to visual tokens, rates the significance of visual tokens within the self-attention matrix corresponding to these text tokens, and incrementally prunes irrelevant visual tokens from the hidden states.
However, FasterVLM \cite{zhang2024fastervlm} highlights a limitation in existing methods, noting that text-to-image attention weights can be biased and may not accurately reflect the importance of visual tokens. Conversely, visual saliency extracted from the visual encoder can mitigate this bias and serve as an effective metric for visual token pruning.

\begin{figure}
    \centering
    \includegraphics[width=1.\linewidth]{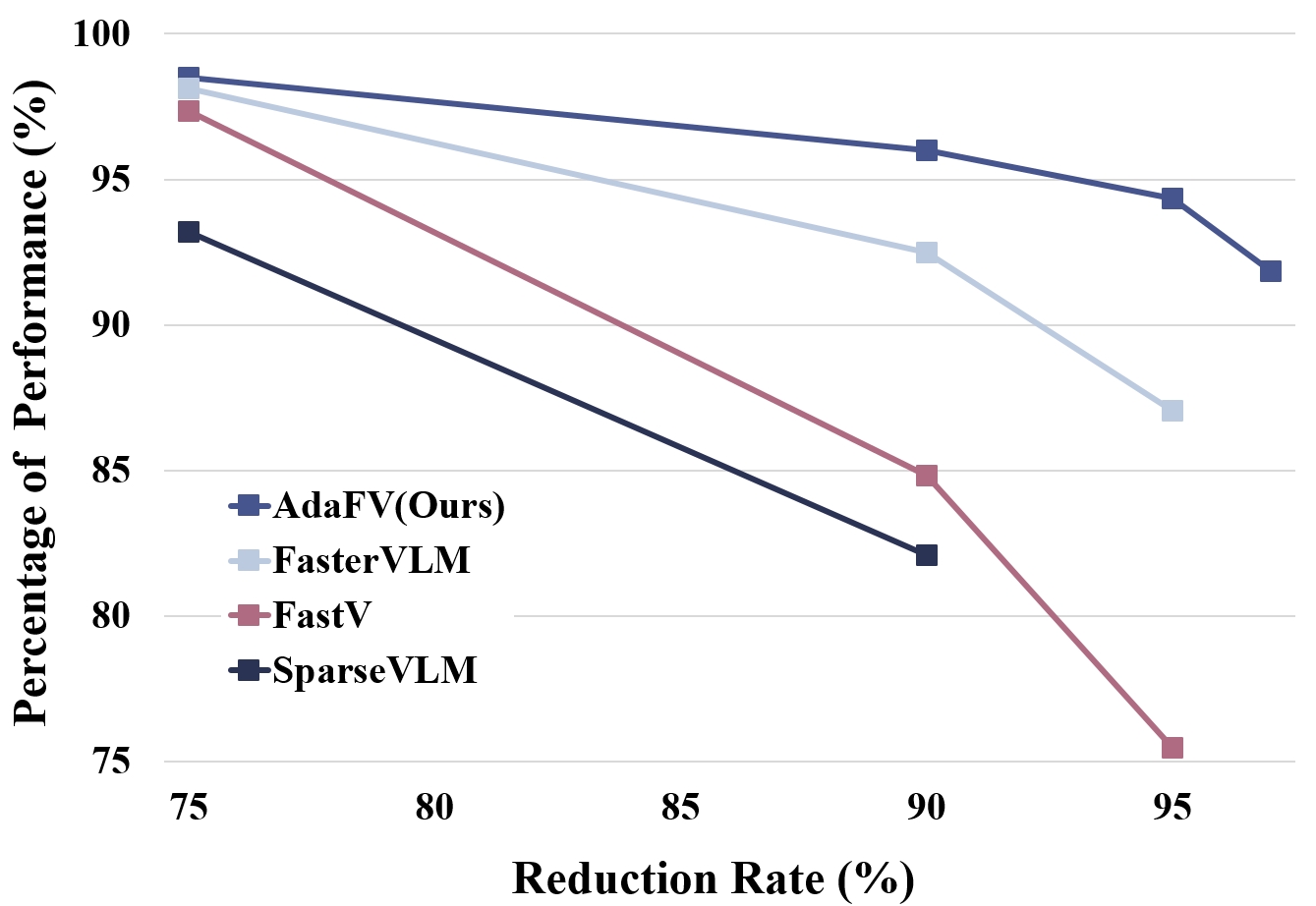}
    \caption{The performance of different training-free VLM acceleration methods on LLaVA-NEXT-7B. AdaFV significantly increased the robustness of the model, especially for a large reduction rate.}
    \label{fig:fig1}
\end{figure}

In this work, we propose leveraging the effectiveness of text-to-image (T2I) similarity and visual saliency for visual token pruning. First, we analyze the alignment of visual tokens with text embeddings in VLMs. Although FasterVLM \cite{zhang2024fastervlm} reveals that self-attention is biased towards less important visual tokens, leading to unreliable token selection, we find that the similarity between text embeddings and visual tokens does not suffer from such misalignment. This similarity can effectively identify prompt-related visual tokens with a reduced token budget. However, certain visual cues, such as those in document images, may not be suitable for cross-modality alignment.
To address these scenarios, we propose a novel self-adaptive cross-modality attention mixture (SACMAM) approach that effectively balances the weight of T2I similarity and visual saliency. To ensure co-dependence on T2I similarity and visual saliency, we measure the overall significance of the selected subset of visual tokens using an independent geometric average and employ a one-step optimization to determine the optimal collection.
Extensive experiments demonstrate that the proposed method achieves state-of-the-art performance on multiple benchmark datasets and is even comparable to fine-tuning methods such as VisionZip \cite{yang2024visionzip}. Our main contributions are as follows:
\begin{enumerate}
    \item We analyze the patterns in multiple VLM benchmark datasets and explain why text-agnostic visual token pruning methods are effective. We also identify scenarios where these methods are sub-optimal, limiting model performance.
    \item We propose a novel self-adaptive cross-modality attention mixture mechanism (SACMAM) that adaptively determines the dominance of T2I attention and visual saliency for visual token pruning, effectively combining the strengths of both metrics.
    \item Extensive experiments show that our approach achieves state-of-the-art performance and is comparable to fine-tuned methods.
\end{enumerate}

\section{Related work}
\subsection{Vision language models (VLMs)}
Significant progress has been made in the development of vision-language models (VLMs). LLaVA \cite{Li-hallucination-2023} was the first approach to effectively combine large language models (LLMs) with foundational vision models. Following LLaVA, a series of VLMs have been proposed, enhancing multi-modal capabilities \cite{lin2023vila, Qwen-VL, chen2024internvl}. The initial models in the LLaVA family only utilized a single image as input, resulting in 576 visual tokens for a 336×336 image. However, this approach often led to significant information loss, thereby degrading model performance. To address this issue, many VLMs have introduced dynamic high-resolution strategies. Despite their success, these high-resolution strategies have significantly increased the number of visual tokens. For instance, LLaVA-NEXT models \cite{liu2024llavanext} involve up to 2880 visual tokens for the same task.
\subsection{VLM acceleration with token pruning}
Token pruning is a straightforward solution for accelerating transformer models. By merging information into a smaller number of tokens, redundant tokens can be pruned to safely accelerate transformer models, with minimal impact on performance while significantly reducing computational costs \cite{kim2022learned, fu2024lazyllm, wang2024zero}. Some approaches have adapted this concept to accelerate VLMs. For example, \citet{chen2024image, ye2024fit} propose measuring the significance of visual tokens based on self-attention extracted from within LLMs. To further accelerate models, some approaches perform token pruning before the encoder layers of LLMs. FasterVLM, for instance, suggests that text-to-image attentions in LLM layers are biased, leading to inaccuracies, and instead proposes using the visual saliency of visual tokens as a metric to prune non-salient tokens. To enhance pruning effectiveness, some approaches also fine-tune VLMs. For example, VisionZip \cite{yang2024visionzip} fine-tunes the MLP projector to improve alignment for text-to-image attention.

\section{Evaluating alignment and similarity in vision-language tasks}

\subsection{Are VQA tasks aligned with visual salient information?}
\label{sec:3.1}

Recent studies indicate that preserving visual tokens that significantly contribute to the [CLS] token enhances performance in visual question-answering (VQA) tasks \cite{zhang2024fastervlm,yang2024visionzip}. This suggests that current VQA tasks are closely related to salient visual information. To investigate this hypothesis, we conducted experiments on five benchmark datasets: MME, MM-Vet, TextVQA, POPE, and GQA. We employed the SAM-2 model \cite{ravi2024sam2} to segment objects based on text prompts and used the CLIP vision encoder \cite{radford2021learning} to generate visual saliency for each image token. The area under the curve (AUC) was calculated to assess the consistency between visual saliency and text-grounded segmentation. Detailed experimental procedures are provided in the Appendix.

As illustrated in Fig.\ref{fig:auc_hist} (a), the average AUC across datasets exceeds 0.5, indicating that VQA tasks are generally grounded in salient visual cues. However, as shown in Fig.\ref{fig:auc_hist} (b–f), numerous tasks are not aligned with salient visual information, highlighting areas for further exploration.
\begin{figure}[htbp]
    \centering
    \includegraphics[width=1.\linewidth]{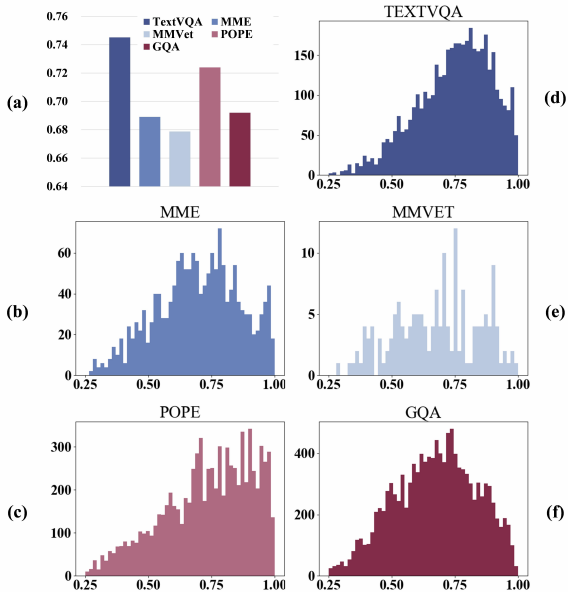}
    \caption{The average AUC on different datasets (a) and the distribution of AUC on each dataset (b$\sim$f).}
    \label{fig:auc_hist}
\end{figure}

\subsection{Text-to-image similarity distribution in the pre-LLM layers}
FasterVLM provides a comprehensive evaluation of text-to-image attention in vision-language models (VLMs), revealing that attention shifts occur in the large language model (LLM) layers. This shift results in later visual tokens receiving more attention than earlier ones, reducing the effectiveness of token selection based on text-to-image attention. However, such attention shifts are absent in CLIP \cite{radford2021learning} and in open-vocabulary detection/segmentation tasks \cite{zheng2024training,bai2024self}. We hypothesize that the attention layers within the LLMs are responsible for this shift, while the original input embeddings—including system, visual, and text embeddings—do not exhibit this behavior.

\begin{figure}[tbp]
    \centering
    \includegraphics[width=1.\linewidth]{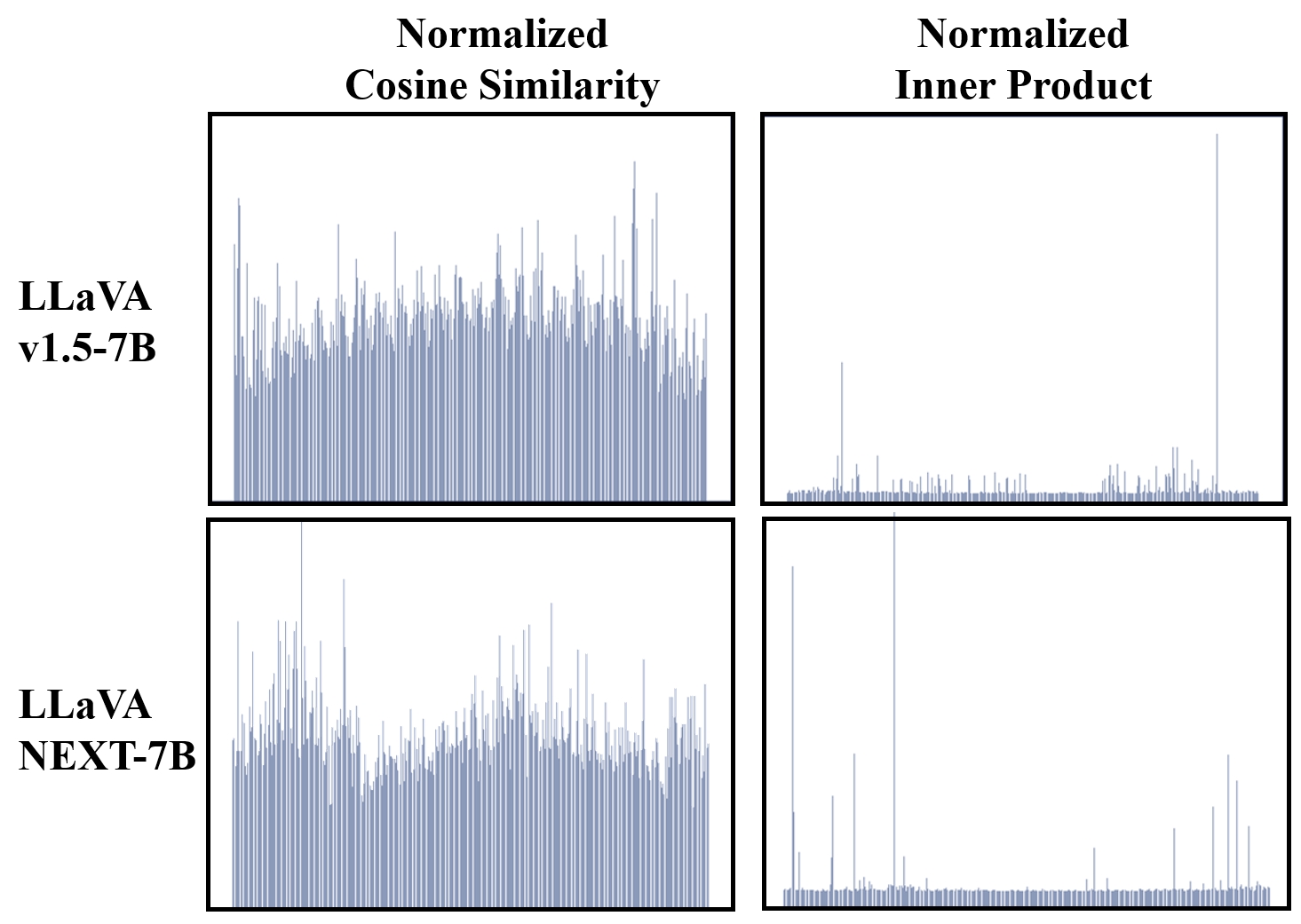}
    \caption{Text-to-image similarity distribution of LLaVA-v1.5-7B and LLaVA-NEXT-vicuna-7B.}
    \label{fig:T2I_attn}
\end{figure}

To test this hypothesis, we conducted an experiment using a subset of the llava dataset \cite{liu2023llava}, measuring the distribution of text-to-image similarity for embedded text tokens and visual embeddings. We employed two metrics to assess alignment: normalized cosine similarity and normalized inner product. The results, visualized in Fig.~\ref{fig:T2I_attn}, show no attention shift in the pre-LLM layers.
However, the normalized inner product produced significant outliers, potentially degrading model performance. In contrast, the normalized cosine similarity demonstrated a more uniform distribution, effectively mitigating outliers and offering greater reliability.

\subsection{Are text embeddings in pre-LLM layers aligned with visual embeddings?}

Having established that text-to-image similarity in pre-LLM layers does not exhibit attention shifts, we next investigate whether text embeddings are sufficiently aligned with visual embeddings to facilitate effective visual token selection. To achieve this, we use the minimum number of visual tokens required as our validation metric. Specifically, if the $N_{th}$ token is the first token relevant to the question, $N$ is served as the number.

\begin{figure}
    \centering
    \includegraphics[width=0.9\linewidth]{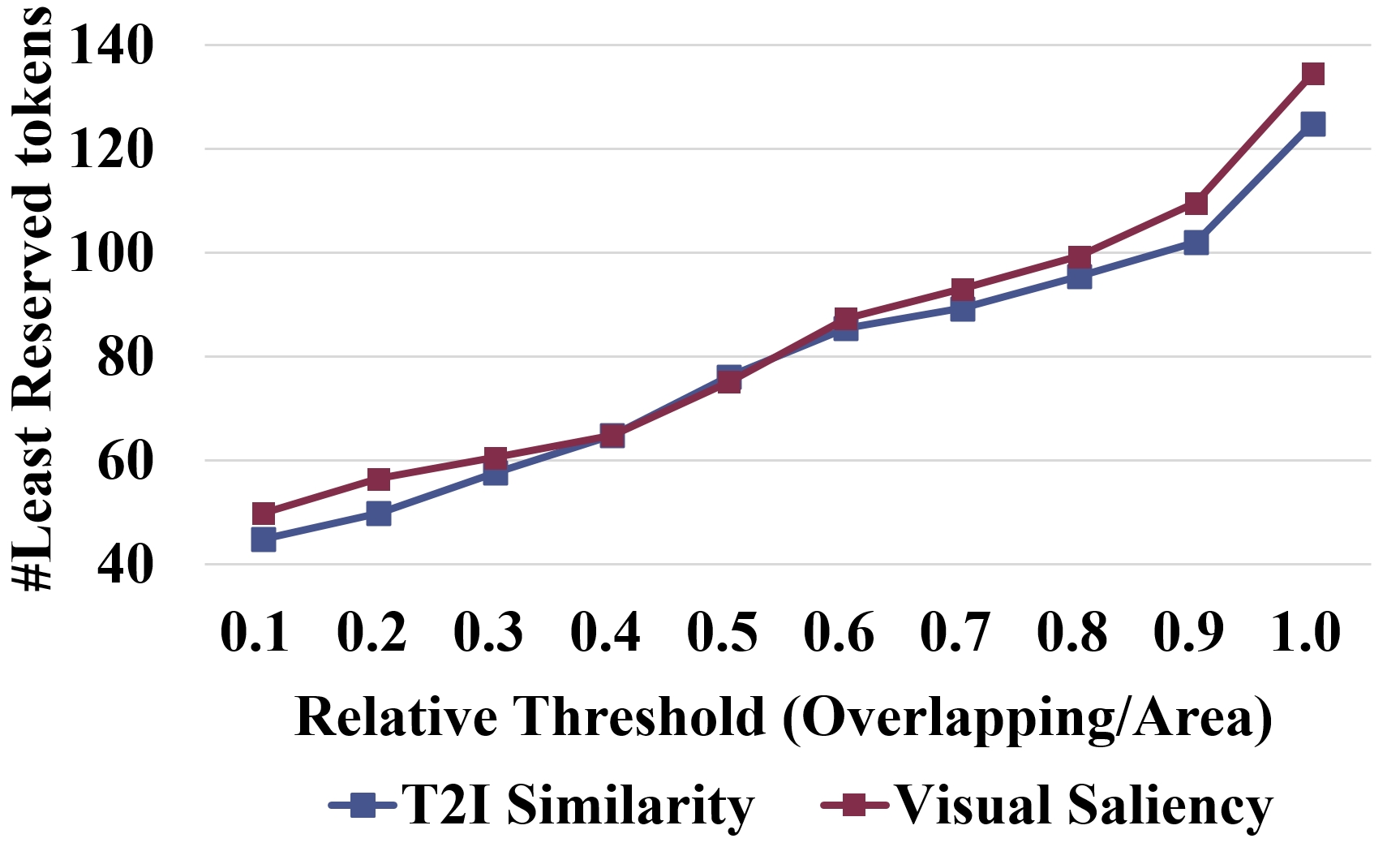}
    \caption{Minimum number of visual tokens to be preserved to select at least one prompt-related visual token, validated on LLaVA-v1.5-7B.}
    \label{fig:least}
\end{figure}

Following the methodology of FasterVLM, we conducted experiments on a subset of the llava data collection. We utilize the same pipeline as described in Sec.~\ref{sec:3.1} to determine the relevant visual tokens. Our findings, shown in Fig.~\ref{fig:least}, indicate that text-to-image similarity requires fewer reserved visual tokens to cover at least one relevant visual token, compared to visual saliency. This demonstrates that text-to-image similarity is an effective metric for visual token preservation.
\section{Method}
\begin{figure*}
    \centering
    \includegraphics[width=.9\linewidth]{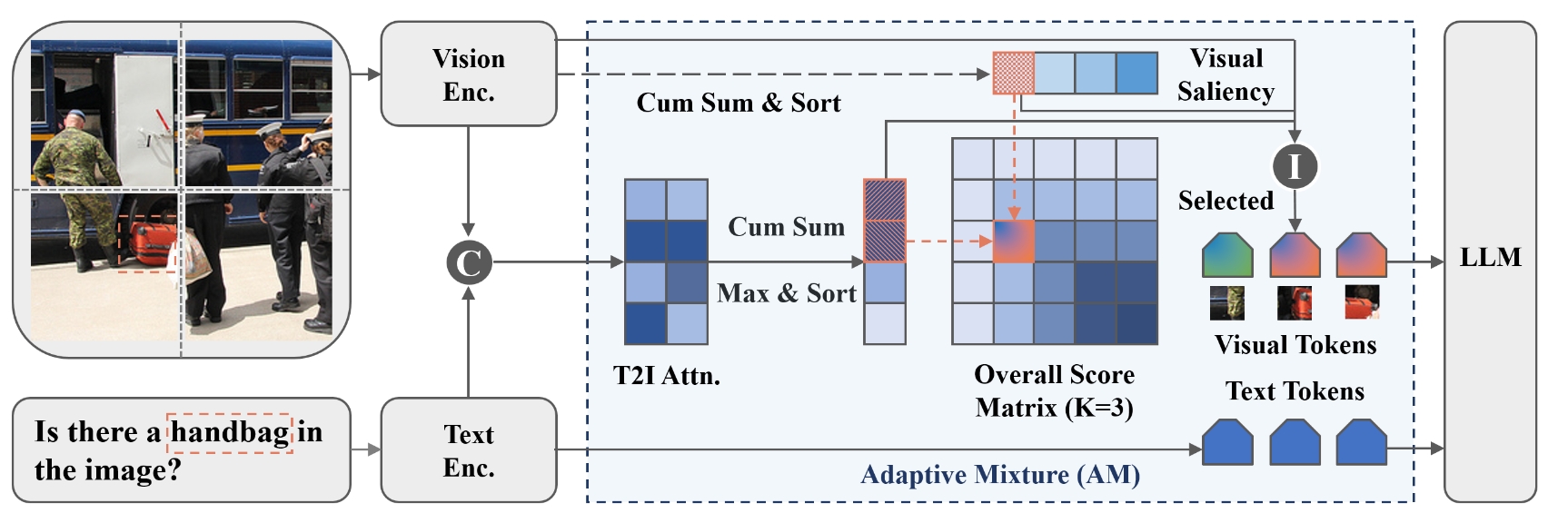}
    \caption{The overall pipeline of the proposed approach. We follow standard VLMs to encode the input image and text. We utilize text-to-image similarity to formulate T2I attention and integrate the attention weights of image tokens by calculating the maximum value. We obtain the overall significance by mixing the T2I attention and visual saliency extracted from the vision encoder and selecting the most informative visual tokens accordingly.}
    \label{fig:overall}
\end{figure*}

\subsection{Fixed cross-modality attention mixture}
As indicated by the aforementioned analysis, relying solely on visual saliency is insufficient to capture all necessary information for addressing questions. This highlights the importance of text-image attention for retaining information. To address this, we leverage both visual saliency and text-image attention for token pruning. A straightforward approach is to select the top-\(K\) tokens based on both visual saliency and text-image similarity.

Formally, let the input text embeddings be \(T_E \in \mathbb{R}^{N_T \times D}\), visual embeddings be \(T_V \in \mathbb{R}^{N_{\text{img}} \times N_I \times D}\), and the [CLS] token be \(C \in \mathbb{R}^{N_{\text{img}} \times D}\), where \(N_T\), \(N_{\text{img}}\), and \(N_I\) represent the number of text tokens, images, and visual tokens per image, respectively. As noted in ViT \cite{dosovitskiy2020image}, the [CLS] token is used for image classification and encapsulates global information. Thus, we use the attention between the [CLS] token and visual embeddings to determine visual saliency, as the attention scale reflects each visual embedding's contribution to global information. The visual saliency is calculated as follows:
\begin{equation}
    \begin{array}{l}
        S^C_i = \text{Softmax}\Big(\frac{C_iW_Q\big((T_V)_iW_K\big)^T}{\lambda}\Big).
    \end{array}
\end{equation}
The similarity between text and images is calculated as:
\begin{equation}
    \begin{array}{l}
        S^{T2I}_{i,j} = \mathop{\text{max}}\limits_k\Big(\frac{(T_E)_k(T_V)_{i,j}^T}{|(T_E)_k||((T_V)_{i,j})|}\Big),
    \end{array}
\end{equation}

We then sample the visual tokens as follows:
\begin{equation}
\left\{
    \begin{array}{l}
        I_i = \Big[\text{argtop-K}\big(S^C_{i,:}\big);\text{argtop-K}\big(S^{T2I}_{i,:}\big)\Big]\\
        (\hat{T}_V)_i=(T_V)_{I_i}.
    \end{array},
\right.
\end{equation}

\subsection{Self-adaptive cross-modality attention mixture (SACMAM)}
Although the fixed approach leverages both global and text-related information, simply combining them can be sub-optimal. To improve token preservation, we propose an adaptive allocation of the token budget based on both metrics.
We employ cosine similarity to select visual tokens aligned with the text. However, the distribution of cosine similarity is not directly comparable to visual saliency which is extracted from self-attention, limiting the effectiveness of mixed token selection. To address this, we introduce a temperature parameter $\tau$ to re-weight the similarity scores, allowing for a balanced comparison. This can be formally expressed as follows:
\begin{equation}
    \begin{array}{l}
        \tilde{S}^{T2I}_{i,j} = \mathop{\text{max}}\limits_k\left(\text{Softmax}\Big(\frac{(T_E)_k(T_V)_{i,j}^T}{|(T_E)_k||((T_V)_{i,j})|\tau}\Big)\right),
    \end{array}
\end{equation}
After re-weighting, the text-image similarity becomes comparable to the visual saliency. For simplicity, we merge the dimensionality of images and tokens per image. Thus, both \(S^C\) and \(\tilde{S}^{T2I}\) are reshaped to \((N_{\text{img}} \times N_I, 1)\).

To select the most informative visual tokens within a budget of \(K\) tokens, a straightforward approach is to choose the top-\(K\) tokens based on their combined scores. This is done as follows:
\begin{equation}
\left\{
    \begin{array}{l}
        \hat{I} = \text{top-K}([S^C,\tilde{S}^{T2I}])\\
        \hat{T}_V=(T_{V})_{\hat{I}}
    \end{array},
\right.
\end{equation}
which is equivalent to optimizing the following objective:
\begin{equation}
    \begin{array}{l}
       \sum\limits_{i\in\mathcal{I}}\tilde{S}^{T2I}_i+\sum\limits_{j\in\mathcal{J}}S^{C}_j, \quad s.t.\quad |\mathcal{I}|+|\mathcal{J}| = K.
    \end{array}
\end{equation}
However, since the distributions of text-image similarity and the visual saliency are different, and only a small group of tokens are selected, selecting tokens depending solely on an individual metric is possible but not expected. To address this, we utilize the geometric mean of the metrics to measure the average importance of the selected tokens:
\begin{equation}
    \begin{array}{l}
        \sqrt{\sum\limits_{i\in\mathcal{I}} \tilde{S}^{T2I}_i\sum\limits_{j\in\mathcal{J}} S^{C}_j},\quad s.t.\quad |\mathcal{I}|+|\mathcal{J}| = K,
    \end{array}
\end{equation}
in which \(\mathcal{I}\) and \(\mathcal{J}\) represent the sets of selected visual token indices based on text-image similarity and visual saliency, respectively.
We start by sorting \(\tilde{S}^{T2I}\) and \(S^C\), resulting in the sorted scores \(\hat{S}^{T2I}\) and \(\hat{S}^C\).
Next, we calculate the cumulative summations of these sorted scores, denoted as \(\mathbf{a}\) and \(\mathbf{b}\), respectively. This is done as follows:
\begin{equation}
    \begin{array}{l}
        \mathbf{a}_0 = 0, \mathbf{b}_0 = 0,\\
        \mathbf{a}_i = \sum\limits_{j=0}^{i-1}\hat{S}^{T2I}_j,\mathbf{b}_i = \sum\limits_{j=0}^{i-1}\hat{S}^{C}_j,
    \end{array}
\end{equation}
Then we calculate the overall metrics as follows:
\begin{equation}
    \begin{array}{l}
        O=\mathbf{a}\mathbf{b}^T.
    \end{array}
\end{equation}
In order that the invalid indices will not be chosen, we utilize a mask $M$ to set the elements of $O$ corresponding to such indices to zero. Specifically, the mask $M$ could be calculated as follows:
\begin{equation}
M_{i,j}=
\left\{
    \begin{array}{l}
        1, i+j\leq K\\
        0, otherwise
    \end{array}.
\right.
\end{equation}
Then the number of tokens selected by text-to-image similarity and visual saliency could be determined as follows:
\begin{equation}
    \begin{array}{l}
        U,V = \mathop{\text{argmax}}\limits_{i,j}\{ (O\otimes M)_{i,j}\}.
    \end{array}
\end{equation}
After determining the number of tokens selected depending on each significance measurement, we choose the top-valued indices for visual token indexing as follows:
\begin{equation}
\left\{
    \begin{array}{l}
        \mathcal{I} = \{i|\text{rank}(S^{T2I}_i)\leq U\}\\
        \mathcal{J} = \{j|\text{rank}(S^{C}_j)\leq V\}
    \end{array}.
    \right.
\end{equation}
The selected visual tokens could finally be formulated as $\{(T_V)_k\}_{k\in\mathcal{I}\cup\mathcal{J}}$. Note that the selected tokens are sorted according to their original position on the image to maintain the correct spatial relationship.
We allow replicated visual tokens for simplicity.
\section{Experiments}
\subsection{Implementation details}
We validate the proposed approach on the LLaVA-v1.5 and LLaVA-NEXT models. We include widely utilized VLM benchmarks to evaluate our method, including GQA \cite{hudson2019gqa}, SQA \cite{lu2022learn}, MME \cite{fu2024mmecomprehensiveevaluationbenchmark}, MMBench \cite{liu2025mmbench}, MMVet \cite{yu2023mm}, TextVQA \cite{singh2019towards} and Pope \cite{Li-hallucination-2023}. All experiments are conducted on the NVIDIA A100-80G GPU.

\begin{table*}[t]
  \centering
  \caption{Comparison with SOTA approaches on LLaVA-NEXT-7B. $^{\dag}$ means that we report both the perception-only score and the summation of the perception score and the cognition score in parenthesis. $^{\ddag}$ represents the model is fine-tuned.}
  \resizebox{1.\linewidth}{!}{
    \begin{tabular}{lcccccccc}
    \toprule
    \multicolumn{1}{c}{Method} & GQA   & SQA-IMG & TextVQA & POPE  & MME$^{\dag}$   & MMB   & MM-Vet & Average \\
    \midrule
    \multicolumn{1}{c}{\multirow{2}[2]{*}{LLaVA-NEXT-7B}} & \multirow{2}[2]{*}{62.93 } & \multirow{2}[2]{*}{69.66 } & \multirow{2}[2]{*}{59.59 } & \multirow{2}[2]{*}{86.32 } & 1513.78  & \multirow{2}[2]{*}{67.70 } & \multirow{2}[2]{*}{42.60 } & \multirow{2}[2]{*}{100.00\%} \\
          &       &       &       &       & (1842.00)  &       &       &  \\
    \midrule

    \multicolumn{9}{c}{Reduction Rate $\approx$ 75\%} \\
    \midrule
    FastV & 60.38  & 69.81  & 58.39  & 83.09  & 1477.31  & 65.64  & 41.10  & 97.35\% \\
    SparseVLM & 60.88  & 67.48  & 58.08  & 70.99  & 1446.10  & 63.83  & 38.00  & 93.19\% \\
    FaseterVLM & 61.31  & 68.82  & 59.33  & 85.50  & 1480.68  & 67.35  & 40.40  & 98.14\% \\
    \textbf{Ours} & \textbf{62.04 } & \textbf{69.31 } & \textbf{58.37 } & \textbf{87.20 } & \textbf{1509.36 } & \textbf{67.35 } & \textbf{39.70 } & \textbf{98.49\%} \\
    \midrule
    VisionZip & 61.30  & 68.10  & 60.20  & 86.30  & 1702.00  & 66.30  &       & 97.75\% \\
    \textbf{Ours} & \textbf{62.04 } & \textbf{69.31 } & \textbf{58.37 } & \textbf{87.20 } & \textbf{1810.07 } & \textbf{67.35 } &       & \textbf{99.13\%} \\
    \rowcolor[rgb]{ .945,  .937,  .925} VisionZip+FT$^{\ddag}$ & 62.40  & 67.90  & 60.80  & 87.60  & 1778.00  & 65.90  &       & 99.00\% \\
    \midrule
    \multicolumn{9}{c}{Reduction Rate $\approx$ 90\%} \\
    \midrule
    FastV & 55.86  & 69.26  & 55.69  & 71.66  & 1282.86  & 61.60  & 22.70  & 84.81\% \\
    SparseVLM & 56.12  & 68.62  & 51.97  & 63.23  & 1332.22  & 54.47  & 24.70  & 82.08\% \\
    FaseterVLM & 58.12  & 68.12  & 57.57  & 80.00  & 1370.11  & 63.32  & 35.70  & 92.47\% \\
    \textbf{Ours} & \textbf{60.65 } & \textbf{68.57 } & \textbf{57.09 } & \textbf{85.98 } & \textbf{1503.25 } & \textbf{66.32 } & \textbf{36.00 } & \textbf{96.00\%} \\
    \midrule
    VisionZip & 59.30  & 67.30  & 58.90  & 82.10  & 1702.00  & 63.10  &       & 95.07\% \\
    \textbf{Ours} & \textbf{60.65 } & \textbf{68.57 } & \textbf{57.09 } & \textbf{85.98 } & \textbf{1812.89 } & \textbf{66.32 } &       & \textbf{97.77\%} \\
    \rowcolor[rgb]{ .945,  .937,  .925} VisionZip+FT$^{\ddag}$ & 61.00  & 67.50  & 59.30  & 86.20  & 1770.00  & 64.40  &       & 97.40\% \\
    \midrule
    \multicolumn{9}{c}{Reduction Rate $\approx$ 95\%} \\
    \midrule
    FastV & 49.83  & 68.52  & 51.85  & 51.66  & 1079.46  & 54.90  & 21.90  & 75.46\% \\
    FaseterVLM & 54.73  & 68.86  & 55.97  & 72.89  & 1225.96  & 60.48  & 31.90  & 87.06\% \\
    \textbf{Ours} & \textbf{58.53 } & \textbf{68.91 } & \textbf{55.11 } & \textbf{85.25 } & \textbf{1452.91 } & \textbf{65.20 } & \textbf{36.20 } & \textbf{94.35\%} \\
    \midrule
    VisionZip & 55.50  & 68.30  & 56.20  & 74.80  & 1630.00  & 60.10  &       & 90.75\% \\
    \textbf{Ours} & \textbf{58.53 } & \textbf{68.91 } & \textbf{55.11 } & \textbf{85.25 } & \textbf{1736.12 } & \textbf{65.20 } &       & \textbf{95.62\%} \\
    \rowcolor[rgb]{ .945,  .937,  .925} VisionZip+FT$^{\ddag}$ & 58.20  & 67.50  & 57.30  & 83.40  & 1699.00  & 63.90  &       & 94.80\% \\
    \bottomrule
    \end{tabular}%
    }
  \label{tab:main}%
\end{table*}
\subsection{Comparison with SOTA approaches}
We compare our proposed approach with other state-of-the-art, training-free token pruning methods. Due to variations in benchmark datasets, reduction rates, and evaluation metrics across different studies (e.g., VisionZip uses the sum of perception and cognition scores, while FasterVLM focuses solely on perception scores), we present our comparisons in Table~\ref{tab:main} for clarity, specifically for the LLaVA-NEXT-vicuna-7B model. Additionally, Table~\ref{tab:otherVLM} briefly demonstrates the effectiveness of other Vision-Language Models (VLMs), with detailed comparisons available in the Appendix.
Our approach achieves state-of-the-art performance among training-free methods and even surpasses some fine-tuned approaches. It shows remarkable robustness, particularly when preserving less than 10\% of visual tokens. Specifically, for the LLaVA-NEXT-vicuna-7B model with a 95\% reduction rate, our method outperforms other training-free methods by 5.0\% and exceeds the fine-tuned VisionZip by 0.8\%.

\begin{table}[htbp]
  \centering
  \caption{Comparison with SOTA approaches
  }
    \begin{tabular}{lccc}
    \toprule
    \multicolumn{1}{c}{\multirow{2}[2]{*}{Method}} & \multicolumn{3}{c}{Reduction Rate} \\
          & 75\%  & 90\%  & 95\% \\
    \midrule
    \multicolumn{4}{c}{LLaVA-1.5-7B} \\
    \midrule
    FastV & 94.67\% & 86.26\% & 72.48\% \\
    SparseVLM & 93.22\% & 78.87\% & 65.85\% \\
    FaseterVLM & 98.32\% & 92.91\% & 87.76\% \\
    \textbf{Ours} & \textbf{97.83\%} & \textbf{93.59\%} & \textbf{88.32\%} \\
    \midrule
    \multicolumn{4}{c}{LLaVA-NEXT-13B} \\
    \midrule
    FaseterVLM & 97.57\% & 92.79\% & 86.52\% \\
    \textbf{Ours} & \textbf{97.75\%} & \textbf{95.40\%} & \textbf{93.14\%} \\
    \midrule
    \multicolumn{4}{c}{LLaVA-NEXT-34B} \\
    \midrule
    FaseterVLM & / & 89.29\% & 83.90\% \\
    \textbf{Ours} & / & \textbf{91.85\%} & \textbf{88.11\%} \\
    \bottomrule
    \end{tabular}%
  \label{tab:otherVLM}%
\end{table}%

\subsection{Ablation study}
\paragraph{Overall ablation}
We conduct an overall ablation study of the proposed approach. As demonstrated in Tab.~\ref{tab:abl}, the T2I attention significantly boosts the model performance, especially when the number of retained tokens is small. The proposed adaptive mixture could further boost the model by $1.0\%$ for a larger than 90\% reduction rate. 

\begin{table}[htbp]
  \centering
  \caption{Ablation study of main modules on LLaVA-NEXT-vicuna-7B}
  \resizebox{1.\linewidth}{!}{
    \begin{tabular}{l|ccc}
    \toprule
    \multicolumn{1}{c|}{\multirow{2}[2]{*}{Model}} & \multicolumn{3}{c}{\#Tokens} \\
          & 720   & 288   & 144 \\
    \midrule
    Ours  & 98.49\% & 96.00\% & 94.35\% \\
    -Adaptive Mixture & 98.40\% & 94.89\%      & 92.62\% \\
    -T2I Attention & 98.18\% & 92.47\% & 87.06\% \\
    \bottomrule
    \end{tabular}%
    }
  \label{tab:abl}%
\end{table}%

\paragraph{Detailed ablation results on specific dataset}
To further understand the influence of the proposed mechanisms, we validate the model on two datasets: Pope and TextVQA. We validate the method on these datasets because we empirically find that the proposed approach achieves superior performance on the Pope dataset, but is less effective on the TextVQA. The results are demonstrated in Tab.~\ref{tab:pope},~\ref{tab:textvqa}. Solely leverages visual saliency on the Pope dataset resulting in a significant degradation of performance, while the T2I similarity is much more effective in comparison. The simple mixture approach exceeds the visual saliency-only method, while slightly beneath the T2I-only. On the contrary, for the TextVQA dataset, the T2I-only method resulted in obvious performance degradation, and the simple mixture approach also resulted in a performance drop. Compared with the simple mixture, the proposed adaptive mixture significantly reduces the performance degradation on both datasets, demonstrating its robustness across different tasks.

\begin{table}[htbp]
  \centering
  \caption{Ablation study on the Pope dataset. ``SACMAM'', ``T2I'' and ``VS'' demonstrate self-adaptive cross-modality attention mixture, text-to-image similarity and visual saliency, respectively.}
  \resizebox{1.\linewidth}{!}{
    \begin{tabular}{cccccc}
    \toprule
    \multicolumn{1}{c}{\multirow{2}[2]{*}{SACMAM}} & \multirow{2}[2]{*}{T2I} & \multicolumn{1}{c}{\multirow{2}[2]{*}{VS}} & \multicolumn{3}{c}{\#Tokens} \\
          &       &       & 720   & 288   & 144 \\
    \midrule
    $\times$     & $\times$     & \checkmark & 85.50  & 80.00  & 72.89  \\
    $\times$     & \checkmark & $\times$     & 87.11  & 86.21  & 85.12  \\
    $\times$     & \checkmark & \checkmark & 87.07  & 85.52  & 84.04  \\
    \checkmark & \checkmark & \checkmark & 87.20  & 85.98  & 85.25  \\
    \bottomrule
    \end{tabular}%
    }
  \label{tab:pope}%
\end{table}%

\begin{table}[htbp]
  \centering
  \caption{Ablation study on the TextVQA dataset. ``SACMAM'', ``T2I'' and ``VS'' demonstrate self-adaptive cross-modality attention mixture, text-to-image similarity and visual saliency, respectively.}
  \resizebox{1.\linewidth}{!}{
    \begin{tabular}{cccccc}
    \toprule
\multicolumn{1}{c}{\multirow{2}[2]{*}{SACMAM}} & \multirow{2}[2]{*}{T2I} & \multicolumn{1}{c}{\multirow{2}[2]{*}{VS}} & \multicolumn{3}{c}{\#Tokens} \\
          &       &       & 720   & 288   & 144 \\
    \midrule
    $\times$     & $\times$     & \checkmark & 59.33  & 57.57  & 55.97  \\
    $\times$     & \checkmark & $\times$     & 57.47      & 54.43      & 51.57 \\
    $\times$     & \checkmark & \checkmark & 57.27  & 55.76  & 53.36  \\
    \checkmark & \checkmark & \checkmark & 58.37  & 57.09  & 55.11  \\
    \bottomrule
    \end{tabular}%
    }
  \label{tab:textvqa}%
\end{table}%

\begin{figure}[htbp]
    \centering
    \includegraphics[width=.95\linewidth]{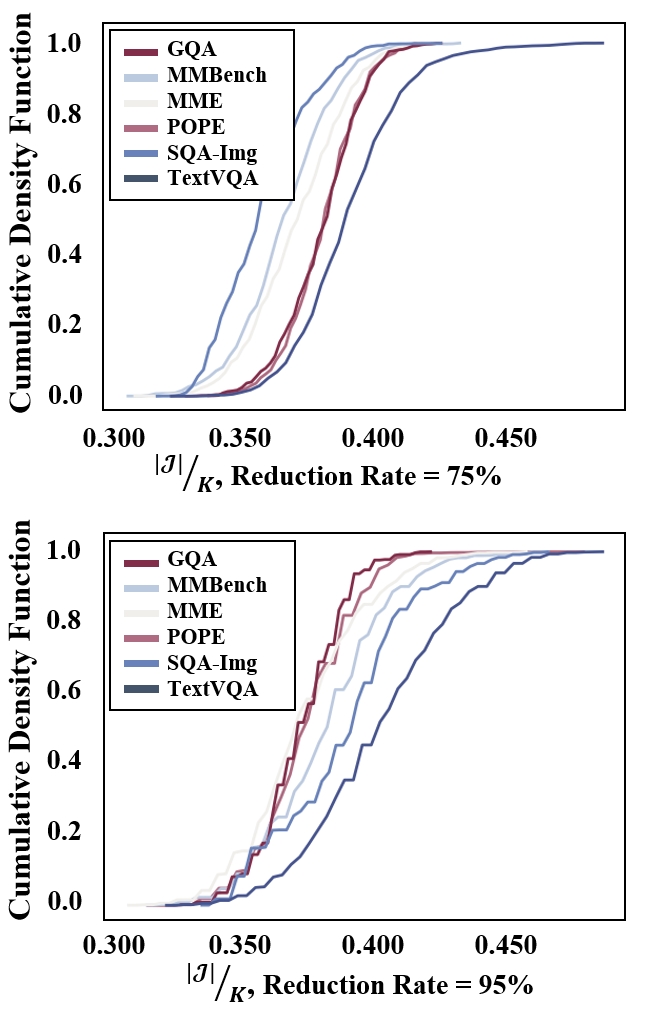}
    \caption{The cumulative density function (CDF) of the proportion of visual saliency-oriented tokens ($\mathcal{J}$) on benchmark datasets of LLaVA-NEXT-7B.}
    \label{fig:attn-dependency}
\end{figure}

\paragraph{Attention dependency analysis}
We further analyze the attention dependency of different datasets and demonstrate the results in Fig.~\ref{fig:attn-dependency}. If the curve is of $\mathcal{I}/K$ left-oriented on the figure, the model is less dependent on the text-to-image similarity. On the contrary, the model is less dependent on visual saliency, if the model relies more on it. We found that the model tends to rely more on visual saliency for tasks that are not related to natural images (TextVQA). This might be because the visual embeddings of the text layouts are not sufficiently aligned with text embeddings, which results in the ineffectiveness of text-to-image similarity in such scenarios. On the contrary, the proposed approach solves the tasks in the POPE dataset, which could be sufficiently solved by text-to-image similarity, depending more on it. This result demonstrates that the proposed self-adaptive cross-modality attention mixture could effectively determine the reliance on visual saliency and text-to-image similarity, boosting the performance of the VLMs.

\begin{figure*}[!htbp]
    \centering
    \includegraphics[width=.8\linewidth]{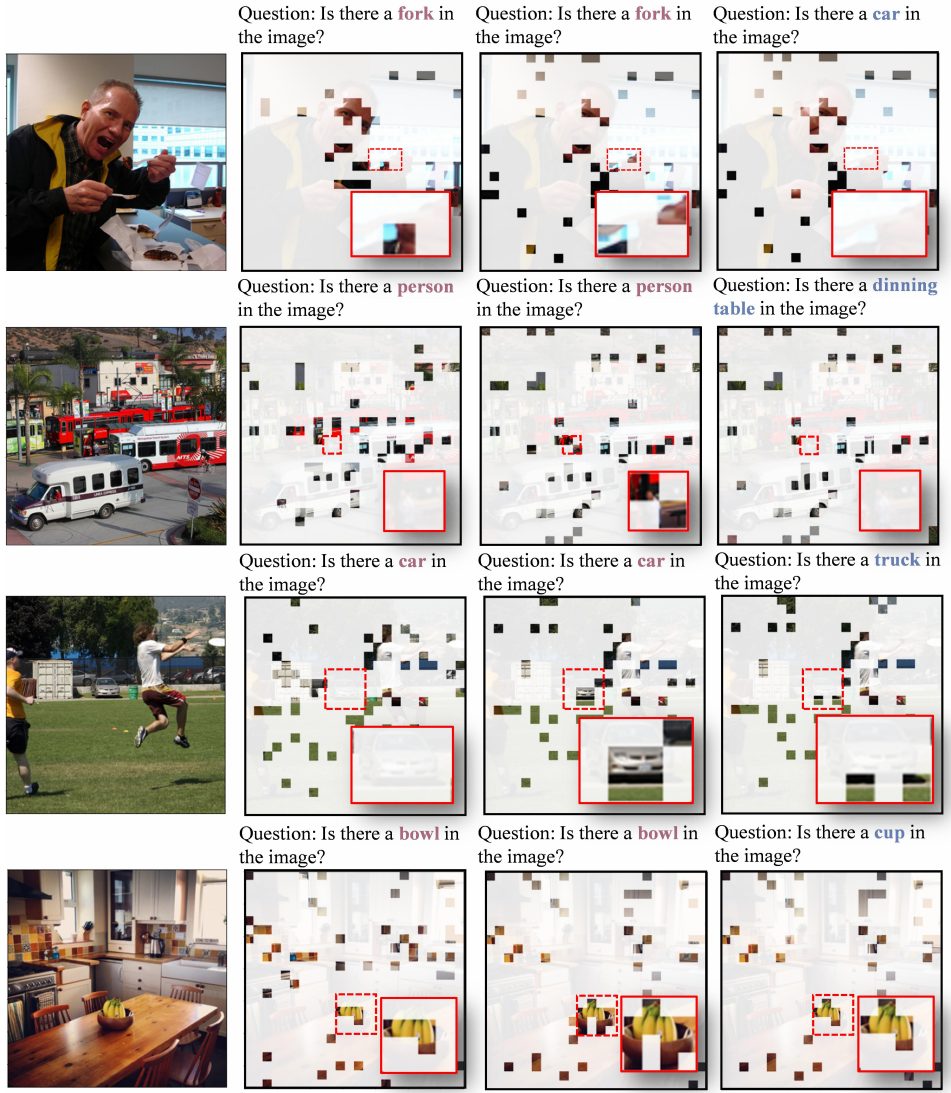}
    \caption{The visualization of selected tokens of FasterVLM (column 2) and the proposed approach (columns 3 and 4). The transparent patches are not selected. Compared with the FasterVLM, the proposed approach managed to focus on question-related tokens. Better zoom in for visualization.}
    \label{fig:tokens}
\end{figure*}
\subsection{Visualization of selected tokens}
We further visualize the selected tokens of the FasterVLM and the proposed approach in Fig.~\ref{fig:tokens}. Since the FasterVLM approach is text-agnostic, the selected visual tokens are consistent with a certain input image, which results in the VLM only accessing the salient objects, and failing to allocate the cases in which the user prompts are about non-salient objects in the image.
On the contrary, the proposed approach effectively leverages the strength of both visual saliency and text-to-image similarity. As depicted in the figure, the proposed approach could draw attention to the detailed information according to the text prompts, and remove it when not relative. 
\section{Conclusion}
In this work, we first analyze the text-to-image similarity in the pre-LLM layers and reveal that the original text embeddings are sufficiently aligned with the visual embeddings for natural image VQA, which suggests that the token pruning before the LLM self-attention is potentially effective. Then we propose an adaptive mixture of visual saliency and text-to-image similarity to select informative visual tokens without introducing additional computational cost. Extensive experiments demonstrate that the proposed approach significantly strengthens the robustness of VLMs for acceleration, and achieves state-of-the-art performance compared with the other approaches.
\section{Limitations}
In this section, we discuss the limitations of the proposed approach. Firstly, although the proposed approach demonstrates the effectiveness of visual token pruning, it relies on sufficient alignment of the text embeddings and visual information, which needs further validation for the native VLMs that utilize VQ-VAE models to generate discrete visual tokens. Second, from visualization, we could still find that there are plenty of redundant visual tokens that are irrelevant to the input prompt. Further exploration of visual information smoothing and filtering is needed to enhance the VLMs' efficiency.

\section*{Acknowledgment}
This project was supported by the supported by Shandong Provincial Natural
Science Foundation (No. ZR2024QF128).

\bibliography{custom}

\appendix
\section{Appendix}
\subsection{Pipeline of analyzing the text prompts and the visual salient information}
To analyze the alignment of text prompts and the visual salient information, we first segment the text-relevant objects with SAM-2 model. To ensure at least one object is included in each image, we gradually decrease the confidence threshold to 0.2 (with a step of 0.01), until at least one segment mask is obtained. If no mask is obtained, we discard the (question, image) pair. Then we utilize CLIP-ViT-L/14 as the vision encoder to extract the contribution of the visual tokens to the [CLS] token. We do not utilize the ViT for ImageNet classification since the CLIP model has a similar nature to the VLMs. The segment mask is separated into $14\times 14$ non-overlap patches to fit the resolution of the vision encoder. A patch is considered to be related to the text prompt if the mask inside the patch occupies more than 50\% of the area of the patch if extra statement is not made. Then a (confidence, label) pair is created for each visual token to calculate the ROC and AUC. We call this the ROC and AUC of visual saliency. Specifically, the confidence is the attention weight, and the label is obtained as follows:
\begin{equation}
    \text{label}=\left\{
    \begin{array}{l}
        0, \text{overlap}<50\% \\
        1, otherwise 
    \end{array}.
    \right.
\end{equation}

\subsection{Effect of benchmark pattern on visual token pruning}
As illustrated in the paper, the average visual saliency AUC of the dataset reflects the pattern of the dataset: whether this dataset tends to contain questions about the visually salient objects of the image. For each dataset, we calculate the relative performance of the FasterVLM and the proposed AdaFV, with a 95\% reduction rate, on different VLMs. The fitted line demonstrates that there is likely to be a pattern: if the AUC of visual saliency is large, depending on the visual saliency is a better choice, otherwise, the model should depend more on text-to-image similarity.
\begin{figure}[htbp]
    \centering
    \includegraphics[width=.95\linewidth]{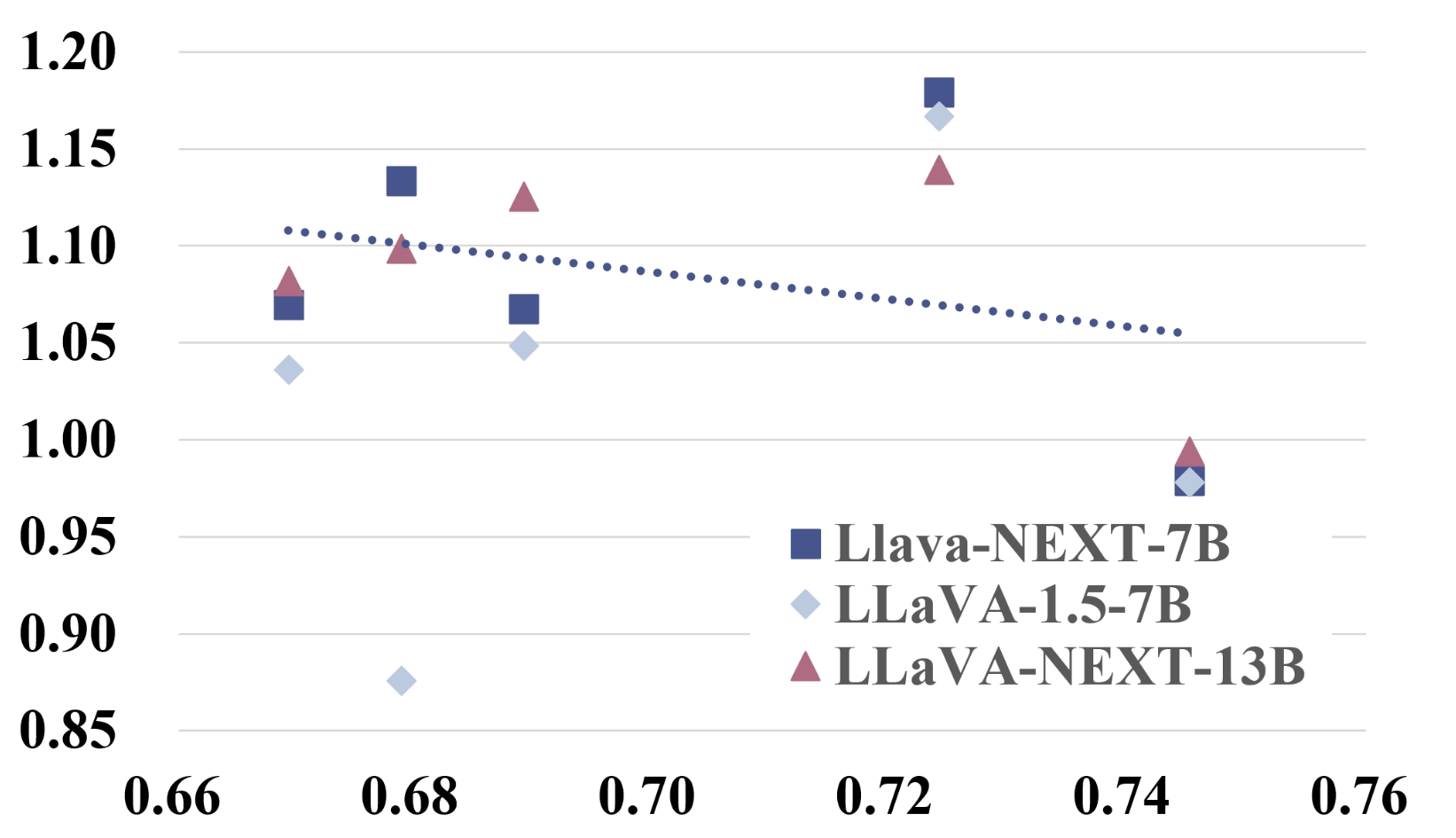}
    \caption{The AUC of visual saliency and text-oriented objects versus the relative performance of AdaFV and FasterVLM.}
    \label{fig:enter-label}
\end{figure}

\subsection{Detailed comparison on LLaVA-v1.5 7B, LLaVA-NEXT 13B and LLaVA-NEXT 34B}
We show a detailed comparison of the token pruning methods on LLaVA-NEXT-13B, LLaVA-NEXT-34B and LLaVA-v1.5-7B in Tab.~\ref{tab:13b}, \ref{tab:34b} and \ref{tab:7b}. 
The proposed AdaFV achieves state-of-the-art performance.

\begin{table*}[htbp]
  \centering
  \caption{Comparison with SOTA approaches on LLaVA-NEXT-13B. $^{\dag}$ means that we report both the perception-only score and the summation of the perception score and the cognition score in parenthesis. $^{\ddag}$ represents the model is fine-tuned.}
  \resizebox{1.\linewidth}{!}{
    \begin{tabular}{lcccccccc}
    \toprule
    \multicolumn{1}{c}{Method} & GQA   & SQA-IMG & TextVQA & POPE  & MME   & MMB   & MM-Vet & Average \\
    \midrule
    \multirow{2}[0]{*}{LLaVA-NEXT-13B} & \multirow{2}[0]{*}{65.40 } & \multirow{2}[0]{*}{73.60 } & \multirow{2}[0]{*}{67.10 } & \multirow{2}[0]{*}{86.20 } & 1575.00  & \multirow{2}[0]{*}{70.00 } & \multirow{2}[0]{*}{48.40 } & \multirow{2}[0]{*}{100.00\%} \\
          &       &       &       &       & (1901.00)  &       &       &  \\

    \midrule
    \multicolumn{9}{c}{Reduction Rate $\approx$ 75\%} \\
    \midrule
    FaseterVLM & 63.05  & 72.88  & 61.67  & 85.27  & 1548.06  & 69.50  & 48.00  & 97.57\% \\
    \textbf{Ours} & \textbf{64.26 } & \textbf{73.33 } & \textbf{61.93 } & \textbf{86.70 } & \textbf{1599.80 } & \textbf{70.10 } & \textbf{44.40 } & \textbf{97.75\%} \\
    \midrule
    VisionZip & 63.00  & 71.20  & 62.20  & 85.70  & 1871.00  & 68.60  &       & 96.93\% \\
    \textbf{Ours} & \textbf{64.26 } & \textbf{73.33 } & \textbf{61.93 } & \textbf{86.70 } & \textbf{1938.72 } & \textbf{70.10 } &       & \textbf{98.82\%} \\
    \rowcolor[rgb]{ .945,  .937,  .925} VisionZip+FT$^{\ddag}$ & 63.70  & 73.20  & 64.40  & 86.30  & 1829.00  & 66.60  &       & 97.38\% \\
    \midrule
    \multicolumn{9}{c}{Reduction Rate $\approx$ 90\%} \\
    \midrule
    FaseterVLM & 59.68  & 71.24  & 60.14  & 80.39  & 1470.98  & 67.61  & 42.90  & 92.79\% \\
    \textbf{Ours} & \textbf{62.78 } & \textbf{73.53 } & \textbf{59.76 } & \textbf{85.79 } & \textbf{1603.05 } & \textbf{69.67 } & \textbf{39.70 } & \textbf{95.40\%} \\
    \midrule
    VisionZip & 60.70  & 70.30  & 60.90  & 82.00  & 1805.00  & 67.20  &       & 94.19\% \\
    \textbf{Ours} & \textbf{62.78 } & \textbf{73.53 } & \textbf{59.76 } & \textbf{85.79 } & \textbf{1912.69 } & \textbf{69.67 } &       & \textbf{97.44\%} \\
    \rowcolor[rgb]{ .945,  .937,  .925} VisionZip+FT$^{\ddag}$ & 62.50  & 72.70  & 63.20  & 85.70  & 1861.00  & 66.90  &       & 96.90\% \\
    \midrule
    \multicolumn{9}{c}{Reduction Rate $\approx$ 95\%} \\
    \midrule
    FaseterVLM & 56.14  & 70.40  & 58.43  & 73.81  & 1388.44  & 64.69  & 34.30  & 86.52\% \\
    \textbf{Ours} & \textbf{60.97 } & \textbf{72.68 } & \textbf{58.05 } & \textbf{84.76 } & \textbf{1557.43 } & \textbf{68.56 } & \textbf{37.90 } & \textbf{93.14\%} \\
    \midrule
    VisionZip & 57.80  & 69.30  & 58.40  & 76.60  & 1739.00  & 64.90  &       & 90.44\% \\
    \textbf{Ours} & \textbf{60.97 } & \textbf{72.68 } & \textbf{58.05 } & \textbf{84.76 } & \textbf{1867.07 } & \textbf{68.56 } &       & \textbf{95.50\%} \\
    \rowcolor[rgb]{ .945,  .937,  .925} VisionZip+FT$^{\ddag}$ & 59.70  & 72.00  & 60.80  & 84.00  & 1766.00  & 65.30  &       & 93.89\% \\
    \bottomrule
    \end{tabular}%
    }
  \label{tab:13b}%
\end{table*}%

\begin{table*}[htbp]
  \centering
  \caption{Comparison with SOTA approaches on LLaVA-NEXT-34B}
  \resizebox{1.\linewidth}{!}{
    \begin{tabular}{lcccccccc}
    \toprule
    \multicolumn{1}{c}{Method} & GQA   & SQA-IMG & TextVQA & POPE  & MME   & MMB   & MM-Vet & Average \\
    \midrule
    \multicolumn{1}{c}{LLaVA-NEXT-34B} & 67.10  & 81.80  & 69.50  & 87.70  & 2028.00  & 79.30  & 57.40  & 100.00\% \\
    \midrule
    \multicolumn{9}{c}{Reduction Rate $\approx$ 90\%} \\
    \midrule
    FaseterVLM & 59.60  & 78.43  & 60.93  & 80.35  & 1869.73  & 75.85  & 42.00  & 89.29\% \\
    \textbf{Ours} & \textbf{62.71 } & \textbf{79.08 } & \textbf{57.92 } & \textbf{86.67 } & \textbf{1958.92 } & \textbf{75.17 } & \textbf{45.50 } & \textbf{91.85\%} \\
    \midrule
    \multicolumn{9}{c}{Reduction Rate $\approx$ 95\%} \\
    \midrule
    FaseterVLM & 55.31  & 78.78  & 58.03  & 74.02  & 1745.38  & 71.64  & 36.90  & 83.90\% \\
    \textbf{Ours} & \textbf{60.12 } & \textbf{78.43 } & \textbf{55.05 } & \textbf{86.44 } & \textbf{1909.81 } & \textbf{74.39 } & \textbf{37.60 } & \textbf{88.11\%} \\
    \bottomrule
    \end{tabular}%
    }
  \label{tab:34b}%
\end{table*}%

\begin{table*}[htbp]
  \centering
  \caption{Comparison with SOTA approaches on LLaVA-v1.5-7B. $^{\dag}$ means that we report both the perception-only score and the summation of the perception score and the cognition score in parenthesis. $^{\ddag}$ represents the model is fine-tuned.}
  \resizebox{1.\linewidth}{!}{
    \begin{tabular}{lcccccccc}
    \toprule
    \multicolumn{1}{c}{Method} & GQA   & SQA-IMG & TextVQA & POPE  & MME   & MMB   & MM-Vet & Average \\
    \midrule
    \multicolumn{1}{c}{\multirow{2}[1]{*}{LLaVA-1.5-7B}} & \multirow{2}[1]{*}{61.94 } & \multirow{2}[1]{*}{69.51 } & \multirow{2}[1]{*}{58.21 } & \multirow{2}[1]{*}{85.88 } & 1506.47  & \multirow{2}[1]{*}{64.69 } & \multirow{2}[1]{*}{31.30 } & \multirow{2}[1]{*}{100.00\%} \\
          &       &       &       &       & (1862.00)  &       &       &  \\

    \midrule
    \multicolumn{9}{c}{Reduction Rate $\approx$ 75\%} \\
    \midrule
    FastV & 56.58  & 69.11  & 57.38  & 73.74  & 1463.39  & 64.00  & 28.60  & 94.67\% \\
    FitPrune & 59.38  & 69.01  & 56.49  & 80.75  & 1472.86  & 63.92  & 28.40  & 96.22\% \\
    SparseVLM & 55.11  & 69.36  & 55.99  & 77.57  & 1351.65  & 59.54  & 29.90  & 93.22\% \\
    FaseterVLM & 58.34  & 67.92  & 57.07  & 83.46  & 1433.76  & 62.54  & 34.20  & 98.32\% \\
    \textbf{Ours} & \textbf{58.38 } & \textbf{69.31 } & \textbf{56.66 } & \textbf{84.72 } & \textbf{1432.68 } & \textbf{62.28 } & \textbf{32.40 } & \textbf{97.83\%} \\
    \midrule
    VisionZip & 57.60  & 68.90  & 56.80  & 83.20  & 1761.70  & 62.00  & 30.00  & 96.12\% \\
    \textbf{Ours} & \textbf{58.38 } & \textbf{69.31 } & \textbf{56.66 } & \textbf{84.72 } & \textbf{1762.32 } & \textbf{62.28 } & \textbf{32.40 } & \textbf{97.77\%} \\
    \rowcolor[rgb]{ .945,  .937,  .925} VisionZip+FT$^{\ddag}$ & 58.90  & 68.30  & 57.00  & 83.70  & 1823.00  & 62.60  & 32.90  & 98.36\% \\
    \midrule
    \multicolumn{9}{c}{Reduction Rate $\approx$ 90\%} \\
    \midrule
    FastV & 51.20  & 69.81  & 54.75  & 57.30  & 1210.36  & 59.97  & 27.20  & 86.26\% \\
    FitPrune & 49.96  & 68.22  & 56.49  & 53.81  & 1147.46  & 56.27  & 21.80  & 81.62\% \\
    SparseVLM & 48.86  & 67.23  & 55.99  & 65.82  & 1030.61  & 49.05  & 18.60  & 78.87\% \\
    FaseterVLM & 54.91  & 68.91  & 55.28  & 75.85  & 1348.63  & 60.57  & 30.10  & 92.91\% \\
    \textbf{Ours} & \textbf{55.30 } & \textbf{68.82 } & \textbf{54.53 } & \textbf{82.33 } & \textbf{1368.28 } & \textbf{60.30 } & \textbf{29.20 } & \textbf{93.59\%} \\
    \midrule
    VisionZip & 55.10  & 69.00  & 55.50  & 77.00  & 1690.00  & 60.10  & 31.70  & 94.02\% \\
    \textbf{Ours} & \textbf{55.30 } & \textbf{68.82 } & \textbf{54.53 } & \textbf{82.33 } & \textbf{1695.42 } & \textbf{60.30 } & \textbf{29.20 } & \textbf{93.63\%} \\
    \rowcolor[rgb]{ .945,  .937,  .925} VisionZip+FT$^{\ddag}$ & 58.90  & 68.80  & 56.00  & 80.90  & 1756.00  & 61.50  & 30.20  & 95.76\% \\
    \midrule
    \multicolumn{9}{c}{Reduction Rate $\approx$ 95\%} \\
    \midrule
    FastV & 46.03  & 70.00  & 51.56  & 35.47  & 971.56  & 50.17  & 18.90  & 72.48\% \\
    FitPrune & 43.60  & 68.32  & 46.75  & 31.17  & 855.21  & 39.69  & 18.00  & 65.85\% \\
    FaseterVLM & 51.51  & 69.56  & 53.09  & 67.24  & 1254.80  & 58.51  & 27.50  & 87.76\% \\
    \textbf{Ours} & \textbf{52.96 } & \textbf{68.42 } & \textbf{51.89 } & \textbf{78.04 } & \textbf{1313.36 } & \textbf{58.51 } & \textbf{24.00 } & \textbf{88.32\%} \\
    \bottomrule
    \end{tabular}%
    }
  \label{tab:7b}%
\end{table*}%

\subsection{Influence of model scale}
\begin{figure*}[htbp]
    \centering
    \includegraphics[width=1.\linewidth]{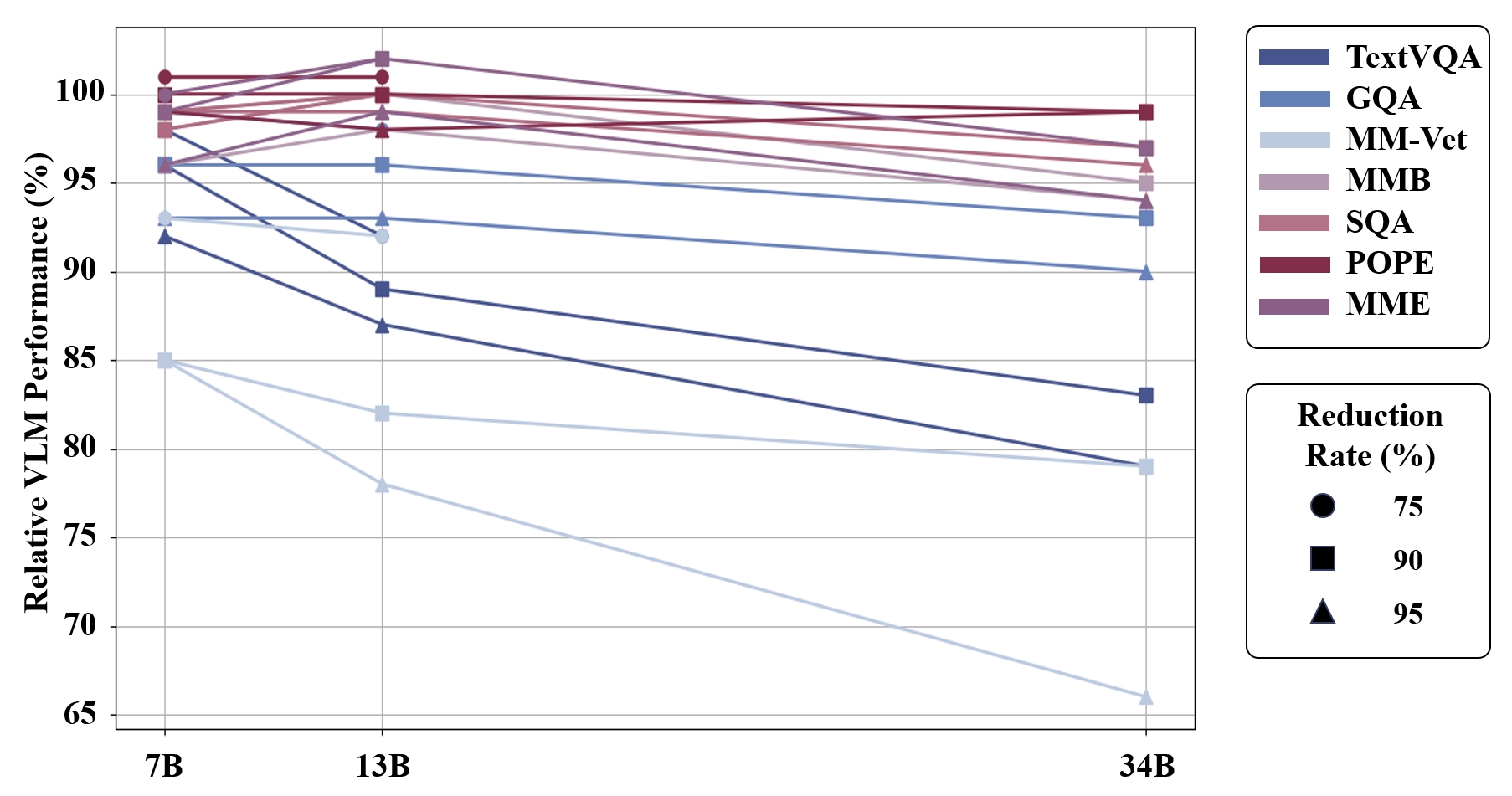}
    \caption{Influence of model scale for visual token pruning across reduction rates and datasets.}
    \label{fig:size}
\end{figure*}

We also visualize the influence of model scale for VLM acceleration. As demonstrated in Fig.~\ref{fig:size}, the increasing scale of the VLMs limits the performance of the visual token pruning in the pre-LLM layers, especially for the text-oriented tasks.

\end{document}